\documentclass[conference]{IEEEtran}
\IEEEoverridecommandlockouts
\usepackage{cite}
\usepackage{amsmath,amssymb,amsfonts}
\usepackage{algorithmic}
\usepackage{graphicx}
\usepackage{textcomp}
\usepackage{xcolor}

\usepackage{booktabs} 
\usepackage{multirow}
\usepackage[normalem]{ulem}
\useunder{\uline}{\ul}{}
\usepackage{enumerate}
\usepackage{threeparttable}

\def\BibTeX{{\rm B\kern-.05em{\sc i\kern-.025em b}\kern-.08em
    T\kern-.1667em\lower.7ex\hbox{E}\kern-.125emX}}
\begin{document}

\title{A Novel Weighted Combination Method for Feature Selection using Fuzzy Sets\\
}

\author{\IEEEauthorblockN{Zixiao Shen, Xin Chen, Jonathan M. Garibaldi}
	\IEEEauthorblockA{\textit{Intelligent Modelling and Analysis Group, School of Computer Science} \\
		\textit{University of Nottingham}, Nottingham, NG8 1BB, United Kingdom \\
		\{Zixiao.Shen, Xin.Chen, Jon.Garibaldi\}@nottingham.ac.uk}
}

\IEEEspecialpapernotice{IEEE International Conference on Fuzzy Systems (FUZZ-IEEE), 2019, New Orleans, USA}

\maketitle

\begin{abstract}
In this paper, we propose a novel weighted combination feature selection method using bootstrap and fuzzy sets. The proposed method mainly consists of three processes, including fuzzy sets generation using bootstrap, weighted combination of fuzzy sets and feature ranking based on defuzzification. We implemented the proposed method by combining four state-of-the-art feature selection methods and evaluated the performance based on three publicly available biomedical datasets using five-fold cross validation. Based on the feature selection results, our proposed method produced comparable (if not better) classification accuracies to the best of the individual feature selection methods for all evaluated datasets. More importantly, we also applied standard deviation and Pearson's correlation to measure the stability of the methods. Remarkably, our combination method achieved significantly higher stability than the four individual methods when variations and size reductions were introduced to the datasets.
\end{abstract}

\section{Introduction}
A fundamental step in many machine learning tasks is feature selection (FS), which helps in dimensionality reduction, data redundant removal, accurate and comprehensive decision makings, etc \cite{foroutan2013new}. Broadly speaking, there are three types of FS methods, i.e. wrapper, embedded and filter methods. Wrapper methods select useful features depending on specified learning algorithms. On account of requiring huge computational resources in learning, wrapper methods become less popular. Embedded methods integrate the FS process into the process of learning such as random forest, etc \cite{liu2009feature}. Contrastively, filter FS methods are independent of learning algorithms which mainly identify a feature subset from the original space based on a set of given evaluation criteria. Due to high computational efficiency, filter methods are widely used for high dimensional datasets \cite{liu2009feature}. Our proposed method is classified as a filter-based method.

For a long time, the research of FS methods mainly focus on improving the classification accuracy and reducing computational time. However, in many practical application areas such as gene selection, biological recognition, cancer detection and etc, a good FS method requires not only high classification accuracy, but also high stability that could be generalized to different datasets \cite{goh2016evaluating, du2017feature}. The stability of a FS method is defined as the robustness of producing consistent feature preferences for independent datasets that are sampled from the same data distribution. It quantifies how the FS outcome is affected by using different datasets from the same application scenario \cite{kalousis2007stability}. With the development of research on high dimensional dataset, the stability of FS becomes increasingly important. In some research areas, the stability of FS results is even more important than the classification performance \cite{chlis2018introducing}.

Improving the stability of FS methods can help us select the relevant features with higher confidence and reduced time consumption of acquiring new data. The study on the stability of FS algorithms can also provide application domain experts with quantified evidence for the reliability of the result using a specific training data. This is particularly crucial in biological applications, e.g. genomics, DNA-microarrays, proteomics and mass spectrometry. Domain experts tend to have less confidence in FS results when slight variations occur in training data. Therefore, the objective of this paper is to develop a FS approach with good classification accuracy and more importantly with high stability.

\section{Literature Review}
Numerous effective techniques have been developed to increase the stability of FS methods. One technique, named \textit{data perturbation strategy}, increases the stability of FS methods by perturbing the training sets, adding new data and integrating multiple FS methods. Specifically, the perturbation method utilizes random sampling to reorganize the original datasets for training. Then it selects and integrates the feature subsets to improve the possibility of selecting a similar feature subset, which leads to higher stability.

Bootstrap aggregation is a meta-algorithm designed to improve the stability and accuracy of machine learning methods. It is also commonly used to improve the stability of FS methods nowadays. Zhou \textit{et al.} \cite{zhou2014stable} have proposed an improved ensemble FS framework based on bootstrap and a multi-ReliefF method. Barbara \textit{et al.} \cite{pes2017exploiting} have designed an ensemble framework with bootstrap sampling for FS. Two strategies (median aggregation and exponential aggregation) are used to integrate the feature ranking results. In bioinformatics area, Saeys \textit{et al.} \cite{saeys2008robust} and Abeel \textit{et al.} \cite{abeel2009robust} have also utilized bootstrap aggregation to generate various bags in order to increase the stability of their FS methods. To increase the robustness of our method, we also use bootstrap to create different subsets of data for FS.

Similar to the case of machine learning algorithms, ensemble techniques can also be used to improve the stability and robustness of FS methods. Due to the constrained search space of a particular feature selector, the optimal feature ranking result may not be achieved. Ensemble FS method can help to alleviate this problem by aggregating the outputs from various feature selectors and produce a better approximation to the optimal feature ranking \cite{saeys2008robust}. On the basis of that, we include an ensemble scheme to our proposed method as well.

Practical biomedical datasets are usually imperfect which contain uncertain texts and incomplete features. The uncertainty will increase after applying bootstrap process. Fuzzy logic techniques are designed to model the vagueness, imprecision and uncertainty \cite{shen2018performance}. In order to better integrate the FS results based on a bootstrap process and an ensemble scheme in a practical setting, it becomes a natural and innovative choice to use fuzzy sets to aggregate the results for FS.

The main contribution of this paper is a novel weighted combination FS method that uses fuzzy sets to aggregate FS results from a bootstrap process and multiple FS methods. The proposed method has been evaluated on three public biomedical datasets for disease classification. The classification accuracies of our method are comparable with the best of the four implemented state-of-the-art methods. More importantly, our method produces significantly higher stability using several evaluation criteria.

\section{Methodology}\label{method}
\begin{figure}[h]
	\centering
	\includegraphics[width=0.5\textwidth]{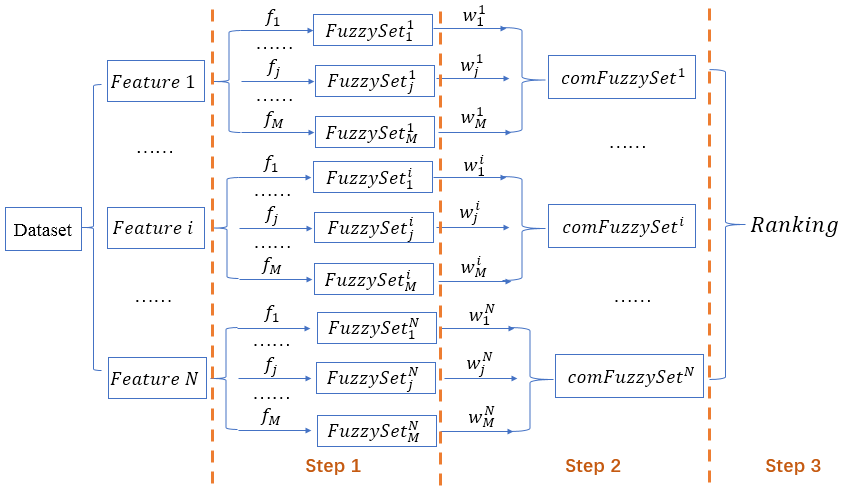}
	\caption[Framework of Fuzzy Sets Generation using Bootstrap Sampling]{Framework of the proposed method}
	\label{framework1}
\end{figure}

By giving a dataset with $S$ data samples and each has $N$ features, our method aims to rank the features from the most to the least significant by combining $M$ different filter-based FS methods. The feature ranking result is then used for classification tasks (e.g. disease discrimination). As illustrated in Fig. \ref{framework1}, the proposed method mainly consists of three steps: (1) fuzzy sets generation using bootstrap; (2) weighted combination of fuzzy sets; (3) defuzzification and feature ranking. The detailed procedures of each step are described in the following subsections.

\subsection{Fuzzy Sets Generation using Bootstrap}
To increase the generalizability of the method, bootstrap is firstly applied to generate a number (denoted as $L$) of data subsets. $M$ different FS methods (denoted as $f_j$, $j \in [1,...,M]$) are then applied to these subsets for obtaining a feature ranking result for each of the subsets per method. Finally, a type 1 fuzzy set is constituted for each feature and for each method based on the $L$ bootstrap subsets. The detailed procedure is described as below. 

\begin{enumerate}
	\item Randomly select $L$ diverse and equal sized datasets with return using bootstrap sampling technique. Each selected dataset is denoted as $Subset \ l$ ($l \in [1,...,L]$).
	\item Implement the FS method $f_j$ on $Subset \ l$ and obtain the feature score (denoted as $Score^i_{lj}$) for the $i$th feature ($i \in [1, ..., N]$).
	\item For each feature, normalize the feature scores $Score^i_{lj}$ between 0 and 1 using min-max normalization \cite{saranya2013study}.
	\item Discretize the feature score, so that all normalized scores belong to the set $U$ which is denoted as $normScore^i_{lj}$.
	\begin{equation}
		U = \{0, 0.01, 0.02,..., 0.98, 0.99, 1\}
	\end{equation}
	\item Combine the normalized feature scores from all $L$ subsets for each of the $i$th feature and the $j$th FS method into a list $F^i_j$.
	\begin{equation}
	F^i_j = \{normScore^i_{lj} | l \in [1,...,L] \}
	\end{equation}
	\item Constitute a type-1 fuzzy set $FuzzySet^i_j$ based on $F^i_j$.
	\begin{equation}\label{eq3}
	FuzzySet^i_j = \{(x, \mu^i_j(x))| x \in U \}
	\end{equation}	
	where the membership function $\mu^i_j(x)$ is represented as:
	\begin{equation}
	\mu^i_j(x)=\left\{
	\begin{array}{rcl}
	freq(x, F^i_j) / L      &      & {x \in   F^i_j}\\
	0     		&      & {x  \notin F^i_j}\\
	\end{array} \right.
	\end{equation}
	$freq(x, F^i_j)$ represents the number of occurrence of $x$ in $F^i_j$.
\end{enumerate}

Finally, the fuzzy set $FuzzySet^i_j$ for the $i$th feature and the $j$th FS method is obtained.

\subsection{Weighted Combination of Fuzzy Sets}
In this step, we combine the results from $M$ different FS methods using weighted summation. Results from various FS methods lead to diversely generated fuzzy sets. Four weighted combination methods are used to integrate these fuzzy sets. For the $i$th feature, the fuzzy sets from different FS methods are combined using weighted sum as described in (\ref{eq6}).

\begin{equation} \label{eq6}
\begin{split}
comFuzzySet^i &= \{(x, \mu^i_{com}(x))| x \in U \} \\
\text{where}   \ \   \mu^i_{com}(x) &= \sum_{j=1}^{M}w^i_j \times \mu^i_j(x) \\
\text{subject to} \    \sum_{j=1}^{M} w^i_j  &= 1
\end{split}
\end{equation}

Four different weight calculation methods are proposed and introduced below.

\subsubsection{\textbf{Equal Weights (EW)}}
This method assigns each fuzzy set of different FS methods with same weights. The definition is shown below.

\begin{equation}
w^i_j = \frac{1}{M}, \  j = 1,..., M
\end{equation}

\subsubsection{\textbf{Reciprocal Standard Deviation Weights (RW)}}
Standard deviation is commonly used to quantify the amount of variation or dispersion of data. We hypothesis that the standard deviation of $F^i_j$ (denoted as $SD(F^i_j)$) is proportional to the uncertainty of the feature score. FS method with high standard deviation has high uncertainty which contributes less for the combined score \cite{hartung2011statistical}. The weight calculation is defined in (\ref{eq9}).

\begin{equation}\label{eq9}
w^i_j = \frac{\frac{1}{SD(F^i_j)}}{\sum_{j=1}^{M}\frac{1}{SD(F^i_j)}}
\end{equation}

\subsubsection{\textbf{One Minus Standard Deviation Weights (OW)}}
Another weight calculation method using standard deviation \cite{hartung2011statistical} is defined in (\ref{eq10}).

\begin{equation}\label{eq10}
w^i_j = \frac{1 - SD(F^i_j)}{\sum_{j=1}^{M}(1 - SD(F^i_j))}
\end{equation}

\subsubsection{\textbf{Matrix Similarity Weights (MW)}}
Arguably, a high certainty does not guarantee a good FS performance. For instance, one method could repetitively produce unreasonable feature ranking result with low standard deviation (high confidence). In this case, the use of standard deviation as above is not an ideal choice for weight calculation. Hence, we propose an alternative weight calculation method based on similarity measurements between the fuzzy set of individual FS method and the combined fuzzy sets of all FS methods. A larger similarity value represents a higher agreement of a particular method with all other methods. The similarity is measured in a 2D matrix (image) form as detailed below.

\begin{itemize}\setlength{\itemsep}{-0.1cm}
	\item Transfer 1-dimensional fuzzy set $FuzzySet^i_j$ (\ref{eq3})  to 2-dimensional binary image $BinaryMatrix^i_j$ (Table \ref{similarityMatrix}) using (\ref{eq12}).\\
		
	\begin{table}[h]
		\centering
		\caption{The definition of $BinaryMatrix^i_j$}
		\begin{tabular}{c|c c c c c c}
			\toprule
			\textbf{Data} & \textbf{0} & \textbf{0.01} & \textbf{...} & \textbf{q} & \textbf{...}   & \textbf{1}\\
			\midrule
			\centering
			$\frac{1}{L}$	 & $s_{1,1}$ &  $s_{1, 2}$  & ...&  $s_{1,q}$  & ... & $s_{1,101}$  \\
			$\frac{2}{L}$	 & $s_{2,1}$ &  $s_{2, 2}$  & ...&  $s_{2, q}$ & ... & $s_{2,101}$  \\
			...			&  ...  &   ...  &  ...  &  ... & ... & ... \\
			$\frac{p}{L}$	 & $s_{p,1}$   &  $s_{p, 2}$  & ...&  $s_{p,q}$  & ... & $s_{p, 101}$\\
			...			&  ...  &   ...  &  ...  &  ... & ... & ...\\
			$\frac{L-1}{L}$	 & $s_{L-1,1}$    &  $s_{L-1, 2}$  & ...&  $s_{L-1,q}$ & ... & $s_{L-1,101}$\\	
			$1$	 & $s_{L,1}$   &  $s_{L,2}$  & ...&  $s_{L,q}$ & ... & $s_{L,101}$  \\			
			\bottomrule
		\end{tabular}
		\label{similarityMatrix}
	\end{table}
	
	\begin{equation}\label{eq12}
	s_{p,q}=\left\{
	\begin{array}{rcl}
	1      &      & { p/L \leq  \mu^i_j(q) }\\
	0     		&      & {others}\\
	\end{array} \right.
	\end{equation}
	
	\item Constitute the combined matrix for the $i$th feature by adding the binary matrices of different FS methods.
	\begin{equation}
	comMatrix^i = \sum_{j=1}^{M} BinaryMatrix^i_j
	\end{equation}
	
	\item Calculate the similarity between $BinaryMatrix^i_j$ and the combined matrix $comMatrix^i$ using (\ref{eq13}).
	\begin{equation}\label{eq13}
	v^i_j = \frac{||BinaryMatrix^i_j \circ comMatrix^i||_1}{||comMatrix^i||_1}
	\end{equation}
	
	The symbol $\circ$ denotes the element-wise product between the two matrices.
	
	\item Normalize the weights using (\ref{sumW}).
	\begin{equation}\label{sumW}
	w^i_j = \frac{v^i_j}{\sum_{j=1}^{M}v^i_j}
	\end{equation}
\end{itemize}

\subsection{Defuzzification and Feature Ranking}
Based on the weighted combination of fuzzy sets, a single value is calculated by a defuzzification process in this step. Center of average defuzzifier is applied here using (\ref{defuzzy}).

\begin{equation}\label{defuzzy}
c_i = \frac{\sum_{x \in U}^{}x \times \mu^i_{com}(x)}{\sum_{x \in U}^{} \mu^i_{com}(x) }
\end{equation}

$c_i$ represents the score of feature $i$. Formally, the result of the proposed FS method can be expressed as $c = (c_1, c_2,..., c_i, ..., c_N)$. According to the final defuzzified feature scores, a feature ranking sequence is obtained by sorting the feature scores from the most to the least significant. The feature ranking is then used as a guidance for subsequent decision making (e.g. disease classification).

\section{Experiments}
The performance of the proposed method has been evaluated and compared for binary disease classification on three public biomedical datasets.

\subsection{Materials}
The three biomedical datasets used in the experiments come from UCI machine learning repository \cite{Dua:2017}. A brief summary of these datasets is given below and listed in Table \ref{descriptionData}.

\begin{table}[h]
	\caption{Description of the biomedical datasets}
	\begin{tabular}{c c c c}
		\toprule
		\textbf{Dataset} & \textbf{No. Features} & \textbf{No. Samples (+/-)}  & \textbf{Samples/Features}\\
		\midrule
		\centering
		PIMA 				  				  			      &  8   &  768 (268/500) 	&  96    \\
		WBC												     &   9  &  699 (241/458) &      77.719\\
		Parkinsons											&  22  &  195 (147/48) &      8.86   \\
		\bottomrule
	\end{tabular}
	\label{descriptionData}
\end{table}

\begin{figure*}[htbp]
	\centering
	\begin{minipage}[t]{0.28\linewidth}
		\centering
		\includegraphics[width=1\textwidth]{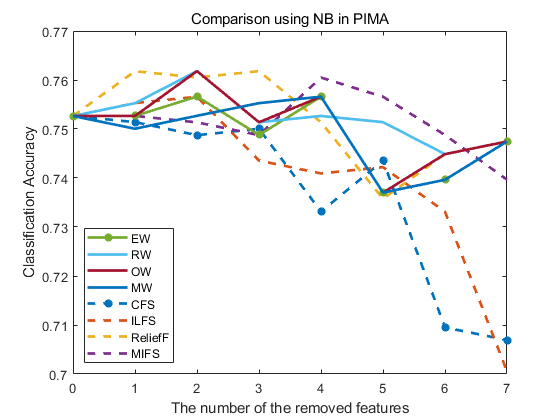}
		\parbox{1cm}{\small \hspace{4.5cm}(a)PIMA}
	\end{minipage}
	\hspace{3ex}   
	\begin{minipage}[t]{0.28\linewidth}
		\centering
		\includegraphics[width=1\textwidth]{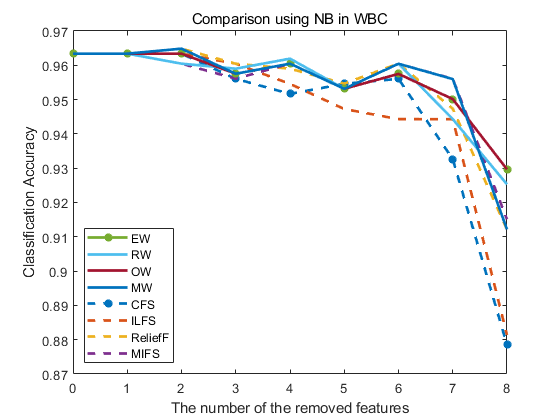}
		\parbox{1cm}{\small \hspace{3.5cm}(b)WBC}
	\end{minipage}
	\hspace{3ex}   
	\begin{minipage}[t]{0.28\linewidth}
		\centering
		\includegraphics[width=1\textwidth]{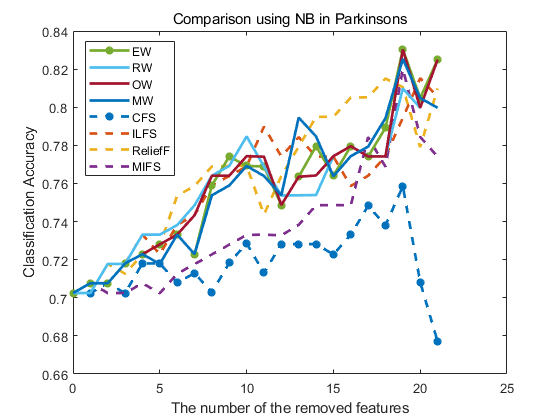}
		\parbox{1cm}{\small \hspace{3.5cm}(c)Parkinsons}
	\end{minipage}
	\caption{Mean classification accuracies using Naive Bayes on different biomedical datasets}
	\label{comNB}
\end{figure*}

\begin{figure*}[htbp]
	\centering
	\begin{minipage}[t]{0.28\linewidth}
		\centering
		\includegraphics[width=1\textwidth]{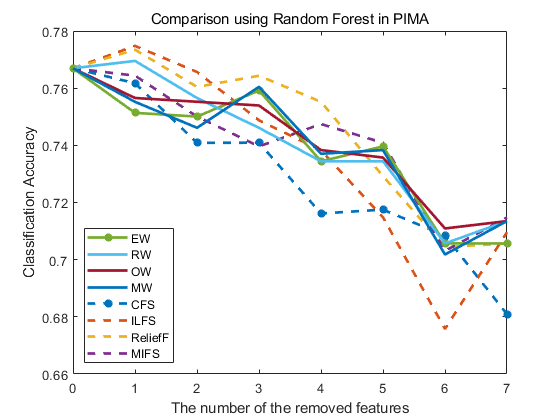}
		\parbox{1cm}{\small \hspace{4.5cm}(a)PIMA}
	\end{minipage}
	\hspace{3ex}   
	\begin{minipage}[t]{0.28\linewidth}
		\centering
		\includegraphics[width=1\textwidth]{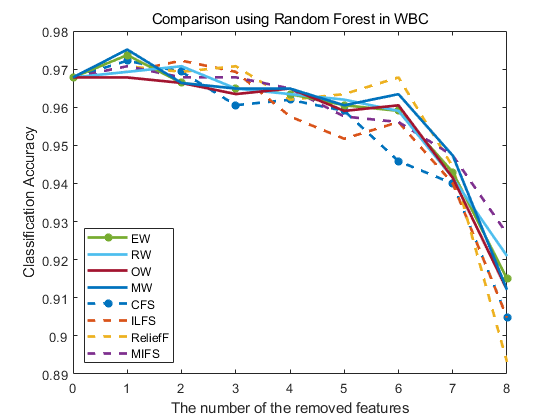}
		\parbox{1cm}{\small \hspace{3.5cm}(b)WBC}
	\end{minipage}
	\hspace{3ex}   
	\begin{minipage}[t]{0.28\linewidth}
		\centering
		\includegraphics[width=1\textwidth]{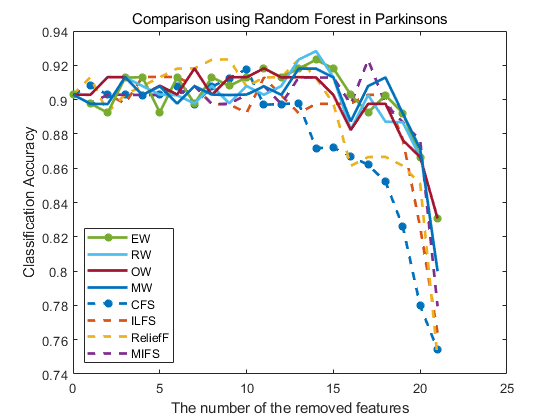}
		\parbox{1cm}{\small \hspace{3.5cm}(c)Parkinsons}
	\end{minipage}
	\caption{Mean classification accuracies using Random Forest on different biomedical datasets}
	\label{comRF}
\end{figure*}

\subsubsection{\textbf{Pima Indians Diabetes Dataset (PIMA)}}
The dataset is originally from National Institute of Diabetes and Digestive and Kidney Diseases. The objective is to predict whether a patient has diabetes based on some diagnostic measurements \cite{smith1988using}.

\subsubsection{\textbf{Wisconsin Breast Cancer (WBC)}}
The dataset was generated by Dr. Wolberg from his clinical cases. For data preprocessing, the sample code ID and the rows with missing values were removed. Then the number of samples became 683 afterwards. Nine visually assessed features were used to predicted benign or malignant of the cancer \cite{wolberg1990multisurface}.

\subsubsection{\textbf{Parkinsons}}
The dataset is composed of a range of biomedical voice measurements from healthy people and the patients with Parkinson's disease (PD). The objective of this data is to discriminate healthy people from those with PD \cite{little2007exploiting}.

\subsubsection{\textbf{Feature Selection Methods}}
Four state-of-the-art filter based FS methods were implemented in the experiments, i.e. Correlation-based Feature Selection (CFS) \cite{hall1999correlation}, Infinite Latent Feature Selection (ILFS) \cite{RoffoICCV17}, ReliefF \cite{robnik2003theoretical} and Mutual Information Feature Selection (MIFS) \cite{zaffalon2002robust}. They were used to generate a combined feature ranking score using our proposed method. The proposed method is generic enough to include larger or smaller number of FS methods.

\begin{table*}[ht]
	\centering
	\caption{Comparison on Achieved Highest Classification Accuracies Between Different FS Methods}
	\begin{threeparttable}
		\begin{tabular}{|c|c|cccc|cccc|}
			\hline
			Datasets                             & Classifiers & \textbf{CFS}          & \textbf{ILFS}         & \textbf{ReliefF}      & \textbf{MIFS}       & \textbf{EW}           & \textbf{RW}           & \textbf{OW}           & \textbf{MW}           \\ \hline
			\multirow{2}{*}{\textit{PIMA}}       & NB          & 0.7527                & 0.7566                &  $0.7618^*$       &  $0.7605’$          & 0.7566                & $0.7618^*$       & $0.7618^*$       & 0.7566                \\ \cline{2-2}
			& RF          & 0.7669                & $0.7748^*$       & $0.7735'$ 			  & 0.7669                & 0.7669                & 0.7695                & 0.7669                & 0.7669                \\ \hline
			\multirow{2}{*}{\textit{WBC}}        & NB          & $0.9634'$   & $0.9634'$   &  $0.9649^*$    &  $0.9634'$  &  $0.9634'$   &  $0.9634'$     &    $0.9634'$  &  $0.9649^*$       \\ \cline{2-2}
			& RF          & 0.9722       & 0.9722       & 0.9708       & 0.9708       & $0.9737'$       & 0.9708       & 0.9678       &  $0.9751^*$       \\ \hline
			\multirow{2}{*}{\textit{Parkinsons}} & NB          & 0.7585                & 0.8152                & 0.8151                & 0.8201                & $0.8303^*$       & 0.8249                & $0.8303^*$       & $0.8253'$ \\ \cline{2-2}
			& RF          & 0.9178       & 0.9132                & 0.9234      &  $0.9232'$                &  $0.9232'$                &  $0.9283^*$                & 0.9182                & 0.9181                \\ \hline
		\end{tabular}
		\begin{tablenotes}
			\footnotesize
			\item[$*$]: shows the best performance;  \hspace{0.2cm}	  $'$: shows the second best performance.
		\end{tablenotes}
	\end{threeparttable}
	\label{comHighest}
\end{table*}

\begin{table*}[ht]
	\centering
	\caption{Comparison of Stability between Different FS methods}
	\begin{threeparttable}
		\begin{tabular}{|c|c|cccc|cccc|}
			\hline
			Index                & Dataset             & \textbf{CFS} & \textbf{ILFS} & \textbf{ReliefF} & \textbf{MIFS} & \textbf{EW}     & \textbf{RW}           & \textbf{OW}           & \textbf{MW}           \\ \hline
			\multirow{4}{*}{ASD} & \textit{PIMA}       & 0.1878       & 0.1248        & 0.0543           & 0.0541        & $0.0397^*$ & 0.0434                &   $0.0393'$   & 0.0443                \\ \cline{2-2}
			& \textit{WBC}        & 0.0873       & 0.0358        & 0.0615           & 0.0453        & $0.0288^*$   & 0.0447                    & $0.0293'$ 			& 0.0315                \\ \cline{2-2}
			& \textit{Parkinsons} & 0.2098       & 0.0533        & 0.0609           & 0.1157        & $0.0336^*$ & 0.0430                &   $0.0341'$    			& 0.0383                \\ \cline{2-10} 
			& Average             & 0.1616       & 0.0713        & 0.0589           & 0.0717        & $0.0340^*$ & 0.0437                &  $0.0343'$ & 0.0380                \\ \hline
			\multirow{4}{*}{APC} & \textit{PIMA}       & 0.6593       & 0.8442        &  $0.9807^*$  & 0.9583        & 0.9573          &  $0.9696'$ & 0.9638                & 0.9482                \\ \cline{2-2}
			& \textit{WBC}        & 0.9391       & 0.9784        & 0.9523           & 0.9829        & $0.9875^*$ & 0.9655                &  $0.9873'$        & 0.9846                \\ \cline{2-2}
			& \textit{Parkinsons} & 0.6031       & 0.8307        & 0.8607           & 0.9111        & 0.9418          & $0.9614^*$       & $0.9453'$	    & 0.9377                \\ \cline{2-10} 
			& Average             & 0.7338       & 0.8844        & 0.9312           & 0.9508        & 0.9622          &   $0.9655^*$       & $0.9655^*$       & $0.9568'$ \\ \hline
		\end{tabular}
		\begin{tablenotes}
			\item[$*$]: shows the best performance;  \hspace{0.2cm}	 $'$: shows the second best performance.
		\end{tablenotes}
	\end{threeparttable}
	\label{asd_apc}	
\end{table*}

\subsection{Evaluation of Classification Accuracy}
FS methods are normally used to improve the classification accuracy with simpler model and reduced number of features. Based on the ranked features, a Naive Bayes (NB) classifier (assuming features are independent to each other) and a non-linear Random Forest (RF) classifier were selected to calculate the classification accuracies by gradually eliminating the least significant features. The implemented filter-based FS methods do not require model training but the NB and RF classifiers are learning based methods. Hence, the experiments were performed using 5-fold cross validation to ensure the training and testing data were independent to each other. The classification accuracy, defined as the number of successfully classified subjects divided by the total number of subjects, was chosen for comparing different methods. The parameter settings are listed as below. $L$ (number of bootstrap subsets) was 100. For each subset, 63.2\% data with return was used. $M$ was 4, indicating four different FS methods (CFS, ILFS, ReliefF and MIFS). Our method of using four different weighting schemes are denoted as EW, RW, OW and MW (see section \ref{method}).

The mean classification accuracies of the five folds using NB classifier and RF classifier by gradually removing the least significant feature are shown in Fig. \ref{comNB} and Fig. \ref{comRF} respectively. In Fig. \ref{comNB} and Fig. \ref{comRF}, our proposed methods produced similar curves as the four competitors. However, our proposed methods achieved the highest classification accuracy when only one significant feature was remained. This indicates a better feature ranking performance using the proposed method. We further list the highest mean classification accuracies of each method in Table \ref{comHighest}. Note that some of the methods produced the same feature ranking sequence, hence the same classification accuracy. Based on the NB classifier, it is observed that our proposed RW and OW methods produced the highest classification accuracy, which was the same as ReliefF for the PIMA data. For the WBC dataset, all methods produced similar classification accuracies with no statistical differences (McNemar's test) between them. For the Parkinsons dataset, our methods (EW, OW) were the top performers but again no statistical difference to the best of the four individual method (i.e. MIFS). The same conclusion can be drawn using the RF classifier. This indicates our proposed combination method is able to achieve comparable (if not better) performance with the best of the individual method.

\subsection{Evaluation of Stability across Different Folds}
As discussed in the introduction section, the stability of FS method is also important. A 'stable' method is able to produce consistent feature ranking results by using different datasets that is sampled from the same data distribution. Hence, in this section, we compare the stabilities of different FS methods based on the feature scoring consistency across the 5 folds. Standard deviation and Pearson's correlation were used as the measurements with detailed descriptions as below. The average standard deviation (ASD) of feature scores is defined in (\ref{ASD}).

\begin{equation}\label{ASD}
ASD = \frac{1}{N} \sum_{i=1}^{N} SD(c_i)
\end{equation}

$SD(c_i)$ stands for the standard deviation of the $i$th feature scores among the 5 folds. ASD is the averaged standard deviation of all $N$ features.

Similar to the work in \cite{kalousis2007stability} and \cite{saeys2008robust}, average Pearson's correlation (APC) was also used for evaluation. APC is defined as the average over all pairwise similarity between the feature scores from different folds derived from the same FS method. The definition is expressed in (\ref{eq20}).

\begin{equation}\label{eq20}
APC = \frac{\sum_{x = 1}^{K} \sum_{y = x +1}^{K} S_P(c^x, c^y)}{K(K-1)/2}
\end{equation}

$S_P(c_x, c_y)$ represents the Pearson's correlation coefficient between the feature scores $c^x$ and $c^y$. $K$ stands for the number of folds ($K = 5$ in our case).

The results of ASD and APC using different FS methods on three biomedical datasets are listed in Table \ref{asd_apc}. Based on the ASD measurement, our proposed method using EW and OW weighting schemes achieved the first and second best performance for all datasets, which were significantly lower than the four individual competitors. Similarly, our RW and OW methods were on average the best two methods based on the APC measurement.  

For the case where ReliefF produced the highest APC in PIMA, its performance on WBC and Parkinsons were low. It is obvious that our proposed method (particularly the OW method) has achieved significantly higher stability than any of the four individual FS method, when the data samples vary across different folds.

\begin{figure*}[htbp]
	\centering
	\begin{minipage}[t]{0.28\linewidth}
		\centering
		\includegraphics[width=1.08\textwidth]{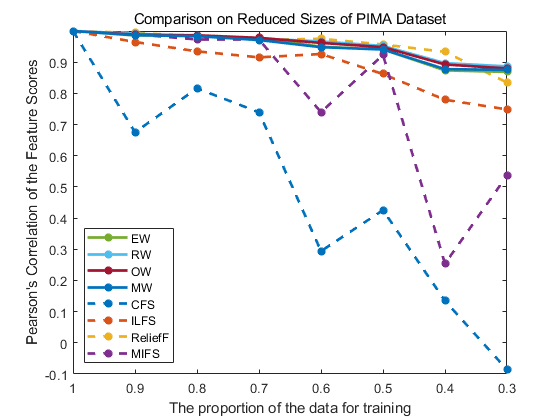}
		\parbox{1cm}{\small \hspace{3.5cm}(a)PIMA}
	\end{minipage}
	\hspace{3ex}   
	\begin{minipage}[t]{0.28\linewidth}
		\centering
		\includegraphics[width=1.08\textwidth]{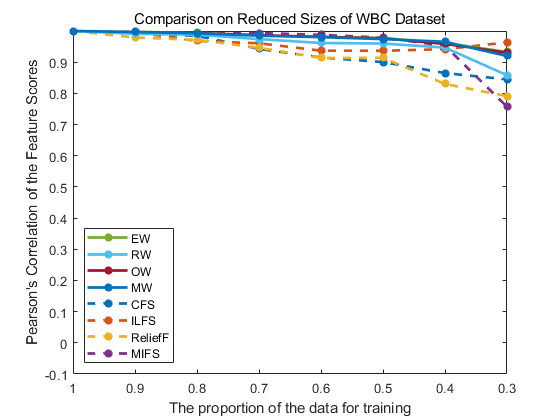}
		\parbox{1cm}{\small \hspace{3.5cm}(b)WBC}
	\end{minipage}
	\hspace{3ex}   
	\begin{minipage}[t]{0.28\linewidth}
		\centering
		\includegraphics[width=1.08\textwidth]{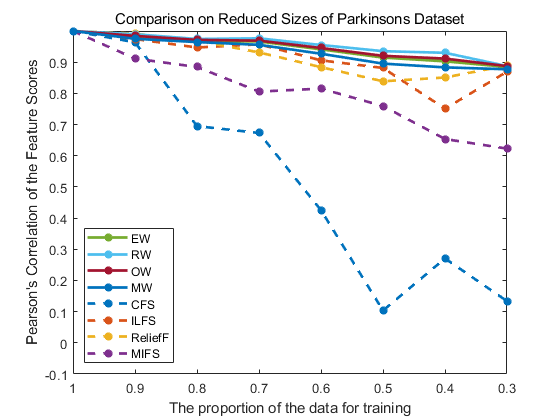}
		\parbox{1cm}{\small \hspace{3.5cm}(c)Parkinsons}
	\end{minipage}
	\caption{Comparison of stability on reduced size of dataset between different FS methods}
	\label{stability}
\end{figure*}

\subsection{Evaluation of Stability on Reduced Size of Dataset}
In this section, we evaluate the stability of FS methods with reduced number of data samples. In practice, a good FS method should be able to produce consistent feature ranking scores even using a subset of the data samples. This is particularly useful when only a small dataset is available.

We repetitively applied the implemented FS methods to the three datasets by gradually reducing the number of data samples. The proportion of the included data is defined by $p$, where $p = 1,0.9, ...,0.3$.  The Pearson's correlation coefficient was calculated between the feature scores using the full size data ($p=1$) and the feature scores using each of the reduced size data ($p = 0.9, ..., 0.3$). With a given $p$, the included data samples were randomly sampled. We repeated each experiment for five times. The mean Pearson's correlation values of 5 experiments by varying $p$ are plotted in Fig. \ref{stability}.

It is seen from Fig. \ref{stability} that with the decreased data size, the mean Pearson's correlation values decrease accordingly for all methods. However, the Pearson's correlation values of our methods using different weight schemes decrease notably slower and maintain steadier than any of the four competitors for all datasets. It is noteworthy to mention that even when $p$ becomes smaller than 0.5, the feature scores of our methods still maintain high similarities ($>0.9$) with the feature scores generated using the full size data. This indicates a superior performance of our method to produce a stable feature ranking result even with significantly reduced size of dataset.

\section{Conclusion}
In this paper, we have proposed a novel weighted combination FS method based on bootstrap and fuzzy sets. By combining four state-of-the-art filter-based FS methods, our method produces an integrated score for feature ranking. We proposed four different weighting schemes to combine different methods. The proposed method has been evaluated on three public biomedical datasets using 5-fold cross validation. The classification accuracy and a number of stability measurements have been used as the evaluation criteria. Our method has produced comparable (if not better) classification accuracies with the best of the individual FS methods for all datasets. Remarkably, our method has achieved significantly higher stability than the four competitors when variations and size reductions are introduced to the dataset. Among four proposed weighting schemes within our method, One Minus Standard Deviation Weight is marginally better than others.

For future work, we will test our proposed method using other weighted combination schemes and defuzzification methods. The proposed method will also be tested on different datasets for various applications with more features and more subjects. The robustness of the proposed method to handle incomplete data and outliers will also be investigated.

\bibliographystyle{IEEEtran}
\bibliography{sample}

\begin{thebibliography}{10}
\providecommand{\url}[1]{#1}
\csname url@samestyle\endcsname
\providecommand{\newblock}{\relax}
\providecommand{\bibinfo}[2]{#2}
\providecommand{\BIBentrySTDinterwordspacing}{\spaceskip=0pt\relax}
\providecommand{\BIBentryALTinterwordstretchfactor}{4}
\providecommand{\BIBentryALTinterwordspacing}{\spaceskip=\fontdimen2\font plus
\BIBentryALTinterwordstretchfactor\fontdimen3\font minus
  \fontdimen4\font\relax}
\providecommand{\BIBforeignlanguage}[2]{{%
\expandafter\ifx\csname l@#1\endcsname\relax
\typeout{** WARNING: IEEEtran.bst: No hyphenation pattern has been}%
\typeout{** loaded for the language `#1'. Using the pattern for}%
\typeout{** the default language instead.}%
\else
\language=\csname l@#1\endcsname
\fi
#2}}
\providecommand{\BIBdecl}{\relax}
\BIBdecl

\bibitem{foroutan2013new}
F.~Foroutan and M.~Eftekhari, ``A new unsupervised fuzzy feature ranking
  measure for feature evaluation,'' in \emph{Fuzzy Systems (IFSC), 2013 13th
  Iranian Conference on}.\hskip 1em plus 0.5em minus 0.4em\relax IEEE, 2013,
  pp. 1--5.

\bibitem{liu2009feature}
H.~Liu, J.~Sun, L.~Liu, and H.~Zhang, ``Feature selection with dynamic mutual
  information,'' \emph{Pattern Recognition}, vol.~42, no.~7, pp. 1330--1339,
  2009.

\bibitem{goh2016evaluating}
W.~W.~B. Goh and L.~Wong, ``Evaluating feature-selection stability in
  next-generation proteomics,'' \emph{Journal of bioinformatics and
  computational biology}, vol.~14, no.~05, p. 1650029, 2016.

\bibitem{du2017feature}
W.~Du, Z.~Cao, T.~Song, Y.~Li, and Y.~Liang, ``A feature selection method based
  on multiple kernel learning with expression profiles of different types,''
  \emph{BioData mining}, vol.~10, no.~1, p.~4, 2017.

\bibitem{kalousis2007stability}
A.~Kalousis, J.~Prados, and M.~Hilario, ``Stability of feature selection
  algorithms: a study on high-dimensional spaces,'' \emph{Knowledge and
  information systems}, vol.~12, no.~1, pp. 95--116, 2007.

\bibitem{chlis2018introducing}
N.-K. Chlis, E.~S. Bei, and M.~Zervakis, ``Introducing a stable bootstrap
  validation framework for reliable genomic signature extraction,''
  \emph{IEEE/ACM Transactions on Computational Biology and Bioinformatics
  (TCBB)}, vol.~15, no.~1, pp. 181--190, 2018.

\bibitem{zhou2014stable}
Q.~Zhou, J.~Ding, Y.~Ning, L.~Luo, and T.~Li, ``Stable feature selection with
  ensembles of multi-relieff,'' in \emph{Natural Computation (ICNC), 2014 10th
  International Conference on}.\hskip 1em plus 0.5em minus 0.4em\relax IEEE,
  2014, pp. 742--747.

\bibitem{pes2017exploiting}
B.~Pes, N.~Dess{\`\i}, and M.~Angioni, ``Exploiting the ensemble paradigm for
  stable feature selection: a case study on high-dimensional genomic data,''
  \emph{Information Fusion}, vol.~35, pp. 132--147, 2017.

\bibitem{saeys2008robust}
Y.~Saeys, T.~Abeel, and Y.~Van~de Peer, ``Robust feature selection using
  ensemble feature selection techniques,'' in \emph{Joint European Conference
  on Machine Learning and Knowledge Discovery in Databases}.\hskip 1em plus
  0.5em minus 0.4em\relax Springer, 2008, pp. 313--325.

\bibitem{abeel2009robust}
T.~Abeel, T.~Helleputte, Y.~Van~de Peer, P.~Dupont, and Y.~Saeys, ``Robust
  biomarker identification for cancer diagnosis with ensemble feature selection
  methods,'' \emph{Bioinformatics}, vol.~26, no.~3, pp. 392--398, 2009.

\bibitem{shen2018performance}
Z.~Shen, X.~Chen, and J.~Garibaldi, ``Performance optimization of a fuzzy
  entropy based feature selection and classification framework,'' in \emph{2018
  IEEE International Conference on Systems, Man, and Cybernetics (SMC)}.\hskip
  1em plus 0.5em minus 0.4em\relax IEEE, 2018, pp. 1361--1367.

\bibitem{saranya2013study}
C.~Saranya and G.~Manikandan, ``A study on normalization techniques for privacy
  preserving data mining,'' \emph{International Journal of Engineering and
  Technology (IJET)}, vol.~5, no.~3, pp. 2701--2704, 2013.

\bibitem{hartung2011statistical}
J.~Hartung, G.~Knapp, and B.~K. Sinha, \emph{Statistical meta-analysis with
  applications}.\hskip 1em plus 0.5em minus 0.4em\relax John Wiley \& Sons,
  2011, vol. 738.

\bibitem{Dua:2017}
\BIBentryALTinterwordspacing
D.~Dheeru and E.~Karra~Taniskidou, ``{UCI} machine learning repository,'' 2017.
  [Online]. Available: \url{http://archive.ics.uci.edu/ml}
\BIBentrySTDinterwordspacing

\bibitem{smith1988using}
J.~W. Smith, J.~Everhart, W.~Dickson, W.~Knowler, and R.~Johannes, ``Using the
  adap learning algorithm to forecast the onset of diabetes mellitus,'' in
  \emph{Proceedings of the Annual Symposium on Computer Application in Medical
  Care}.\hskip 1em plus 0.5em minus 0.4em\relax American Medical Informatics
  Association, 1988, p. 261.

\bibitem{wolberg1990multisurface}
W.~H. Wolberg and O.~L. Mangasarian, ``Multisurface method of pattern
  separation for medical diagnosis applied to breast cytology.''
  \emph{Proceedings of the national academy of sciences}, vol.~87, no.~23, pp.
  9193--9196, 1990.

\bibitem{little2007exploiting}
M.~A. Little, P.~E. McSharry, S.~J. Roberts, D.~A. Costello, and I.~M. Moroz,
  ``Exploiting nonlinear recurrence and fractal scaling properties for voice
  disorder detection,'' \emph{Biomedical engineering online}, vol.~6, no.~1,
  p.~23, 2007.

\bibitem{hall1999correlation}
M.~A. Hall, ``Correlation-based feature selection for machine learning,'' 1999.

\bibitem{RoffoICCV17}
G.~Roffo, S.~Melzi, U.~Castellani, and A.~Vinciarelli, ``Infinite latent
  feature selection: A probabilistic latent graph-based ranking approach,'' in
  \emph{2017 IEEE International Conference on Computer Vision (ICCV)}, Oct
  2017.

\bibitem{robnik2003theoretical}
M.~Robnik-{\v{S}}ikonja and I.~Kononenko, ``Theoretical and empirical analysis
  of relieff and rrelieff,'' \emph{Machine learning}, vol.~53, no. 1-2, pp.
  23--69, 2003.

\bibitem{zaffalon2002robust}
M.~Zaffalon and M.~Hutter, ``Robust feature selection by mutual information
  distributions,'' in \emph{Proceedings of the Eighteenth conference on
  Uncertainty in artificial intelligence}.\hskip 1em plus 0.5em minus
  0.4em\relax Morgan Kaufmann Publishers Inc., 2002, pp. 577--584.

\end{thebibliography}

\end{document}